\documentclass[twoside,11pt]{article}
\usepackage{pgm}

\usepackage[utf8]{inputenc}
\inputencoding{utf8}
\usepackage{soul}
\usepackage{booktabs}
\usepackage[colorlinks,
            linkcolor=blue,
            citecolor=blue,
            urlcolor=magenta,
            linktocpage,
            plainpages=false]{hyperref}
\usepackage[table]{xcolor}

\usepackage{mathtools}
\usepackage{algorithm}
\usepackage{algpseudocode}

\ShortHeadings
  {Using Mixed-Effects Models to Learn Bayesian Networks from Related Data Sets}
  {Scutari, Marquis and Azzimonti}

\newcommand{\mref}[1]{(\ref{#1})}

\newcommand{\Prob}{\operatorname{P}}
\newcommand{\given}{\operatorname{|}}

\newcommand{\X}{\mathbf{X}}
\newcommand{\G}{\mathcal{G}}
\newcommand{\D}{\mathcal{D}}
\newcommand{\Z}{\mathbf{Z}_i}

\newcommand{\PXi}{\Pi_{X_i}}
\newcommand{\bPXi}{\boldsymbol{\Pi}_{X_i}}

\newcommand{\Sigmai}{\boldsymbol{\Sigma}_i}
\newcommand{\tildeSigmai}{\tilde{\boldsymbol{\Sigma}}_i}

\newcommand{\XPi}{X_i \given \PXi}
\newcommand{\T}{\Theta_{X_i}}

\newcommand{\BIC}{\operatorname{BIC}}
\newcommand{\Xij}{X_{ij}}

\renewcommand{\Im}{\mathbf{I}}
\newcommand{\mXi}{\mu_{i}}
\newcommand{\bbXi}{\mathbf{b}_{i}}
\newcommand{\bbij}{\mathbf{b}_{ij}}
\newcommand{\bXi}{\boldsymbol{\beta}_{i}}
\newcommand{\eXi}{\varepsilon_{i}}
\newcommand{\eXij}{\varepsilon_{ij}}
\newcommand{\sXi}{\sigma^2_{i}}
\newcommand{\DXi}{\Delta_{X_i}}
\newcommand{\GXi}{\Gamma_{X_i}}
\newcommand{\bGXi}{\boldsymbol{\Gamma}_{X_i}}
\newcommand{\mXd}{\mu_{ij}}
\newcommand{\bXd}{\boldsymbol{\beta}_{ij}}
\newcommand{\sXd}{\sigma^2_{ij}}

\newcommand{\AP}{\overline{|\PXi|}}
\newcommand{\BT}{\mathcal{B}_\mathrm{TRUE}}
\newcommand{\BG}{\mathcal{B}_\mathrm{GBN}}
\newcommand{\BCG}{\mathcal{B}_\mathrm{CGBN}}
\newcommand{\BLME}{\mathcal{B}_\mathrm{LME}}
\newcommand{\SHD}{\operatorname{SHD}}
\newcommand{\KL}{\operatorname{KL}}
\newcommand{\RMAD}{\operatorname{RMAD}}
\newcommand{\Fone}{F\textsubscript{1}}

\begin{document}

\title{Using Mixed-Effects Models to Learn Bayesian Networks from Related
  Data Sets}

\author{%
   \Name{Marco Scutari \Email{scutari@bnlearn.com} \\
   \addr Istituto Dalle Molle di Studi sull'Intelligenza Artificiale (IDSIA),
     Lugano, Switzerland
   \and
   \Name{Christopher Marquis} \Email{christopher.marquis@epfl.ch} \\
   \addr Ecole Polytechnique Fédérale de Lausanne (EPFL), Lausanne, Switzerland
   \and
   \Name{Laura Azzimonti} \Email{laura.azzimonti@idsia.ch} \\
   \addr Istituto Dalle Molle di Studi sull'Intelligenza Artificiale (IDSIA),
     Lugano, Switzerland}
}

\maketitle

\begin{abstract}

  We commonly assume that data are a homogeneous set of observations when
  learning the structure of Bayesian networks. However, they often comprise
  different data sets that are related but not homogeneous because they have
  been collected in different ways or from different populations.

  In our previous work \citep{ijar21-laura}, we proposed a closed-form Bayesian
  Hierarchical Dirichlet score for discrete data that pools information across
  related data sets to learn a single encompassing network structure, while
  taking into account the differences in their probabilistic structures. In this
  paper, we provide an analogous solution for learning a Bayesian network from
  continuous data using mixed-effects models to pool information across the
  related data sets. We study its structural, parametric, predictive and
  classification accuracy and we show that it outperforms both conditional
  Gaussian Bayesian networks (that do not perform any pooling) and classical
  Gaussian Bayesian networks (that disregard the heterogeneous nature of the
  data). The improvement is marked for low sample sizes and for unbalanced data
  sets.

\end{abstract}
\begin{keywords}
  Gaussian Bayesian networks; conditional Gaussian Bayesian networks;
  random effects models; structure learning.
\end{keywords}

\section{Introduction}

Bayesian networks \citep[BNs; ][]{koller} are a graphical model defined over a
set of random variables $\X = \{X_1, \ldots, X_N\}$, each describing some
quantity of interest, that are associated with the nodes of a directed acyclic
graph (DAG) $\G$. Arcs in $\G$ express direct dependence relationships between
the variables in $\X$, with graphical separation in $\G$ implying conditional
independence in probability. As a result, $\G$ induces the factorisation
\begin{equation}
  \Prob(\X \given \G, \Theta) = \prod\nolimits_{i = 1}^N \Prob(\XPi, \T),
\label{eq:parents}
\end{equation}
in which the joint probability distribution of $\X$ (with parameters $\Theta$)
decomposes in one local distribution for each $X_i$ (with parameters $\T$,
$\bigcup_{i=1}^N \T = \Theta$) conditional on its parents $\PXi$. In this paper
we will assume that $\X$ is a multivariate normal random variable and that the
$X_i$ are univariate normals linked by linear dependencies, thus focusing on the
class of BNs known as \emph{Gaussian BNs} \citep[GBNs;][]{heckerman3}. The
parameters of their local distributions can be equivalently written as the
partial correlations between $X_i$ and each parent given the others or as the
coefficients $\bXi$ of the linear regression model,
\begin{align}
  &X_i = \mXi + \bPXi\bXi + \eXi,& &\eXi \sim
    N(\mathbf{0}, \sXi \Im_n),
\label{eq:gnode}
\end{align}
where $\bPXi$ is the design matrix associated to the parents of $X_i$, $n$ is
the sample size and $\Im_n$ is an $n \times n$ identity matrix, so that
$\T = \{ \mXi, \bXi, \sXi\}$. GBNs are a particular case of conditional Gaussian
BNs \citep[CGBNs;][]{lauritzen}, which assume that $\X$ is a mixture of
multivariate normals and thus can model discrete as well as continuous
variables. Discrete $X_i$ are only allowed to have discrete parents and are
assumed to follow a multinomial distribution with parameters $\T$ organised in
conditional probability tables. Continuous $X_i$ are allowed to have both
discrete and continuous parents (denoted $\DXi$ and $\GXi$ respectively, with
$\DXi \cup \GXi = \PXi$), and their local distributions are mixtures of linear
regressions with one component for each configuration of $\DXi$:
\begin{align}
  &\Xij = \mXd + \bGXi\bXd + \eXij,&
  &\eXij \sim N(\mathbf{0}, \sXd \Im_{n_j}),\
   j = 1, \ldots, |\DXi|, \ \sum\nolimits_j n_j = n.
\label{eq:cgnode}
\end{align}
$\Xij$ refers to $X_i$ for the $j$th configuration of $\DXi$, $\sXd \Im_{n_j}$
is the associated $n_j \times n_j$ diagonal covariance matrix and $\bGXi$ is the
design matrix for the continuous parents $\GXi$. Hence \linebreak
$\T = \bigcup_{j = 1}^{|\DXi|}\{ \mXd, \bXd, \sXd\}$. If $X_i$ has no discrete
parents, \mref{eq:cgnode} simplifies to \mref{eq:gnode}. In fact, CGBNs include
GBNs and discrete BNs \citep{heckerman} as particular cases if $\X$ only
contains continuous or discrete variables respectively.

Learning a BN from a data set $\D$ involves two steps: \emph{structure learning}
and \emph{parameter learning}. Structure learning consists in finding the DAG
$\G$ that encodes the dependence structure of the data and is usually performed
by maximising $\Prob(\G \given \D)$ or some alternative goodness-of-fit measure
(score-based learning). As an alternative, $\G$ can also be estimated by using
conditional independence tests to identify which conditional independence
relationships are supported by the data (constraint-based learning). In the case
of GBNs, structure learning can be implemented either by finding the DAG that
maximises the Bayesian Information Criterion \citep[BIC;][]{schwarz}
\begin{equation}
  \BIC(\G, \Theta \given \D) =
    \sum\nolimits_{i = 1}^{N} \log \Prob(\XPi, \T) - \frac{\log(n)}{2} |\T|
\label{eq:bic}
\end{equation}
or $\Prob(\G \given \D)$, which is called the BGe score, or by testing partial
correlations with Fisher's Z test or the exact Student $t$ test. Both
approaches, as well as the algorithms used to learn $\G$ from $\D$ are reviewed
in \citet{ijar19}. Parameter learning then consists in estimating the parameters
$\Theta$ given the $\G$ obtained from structure learning, typically by means of
maximum likelihood or Bayesian methods. The parameters of a GBN are usually
estimated using the maximum likelihood estimates of regression coefficients from
classical statistics.

All tests and scores in common use in the literature assume that observations
are independent and identically distributed, which implies that $\D$ is
homogeneous. This assumption, however, is easily violated when we pool together
several small data sets, collected under similar but not identical conditions,
in order to achieve a sample size large enough for relevant relationships
to be detectable. Typical examples are data collected within multi-centre
clinical trials \citep{spiegelhalter}, in which differences in both protocols
and patient populations across institutions should be properly modelled, and
within ecology and environmental studies, in which different patterns of
measurement errors and limitations in different environments should be taken
into account \citep{ecology}. The task of efficiently modelling such related
data sets is usually tackled by hierarchical models \citep{gelman}, which pool
the information common to the different subsets of the data while encoding the
information that is specific to each subset. \emph{Linear mixed-effects models}
\citep{pinheiro} are classical statistical methods used to model correlated data
and data with a multilevel or hierarchical structure, such as related data sets.
In this paper, we will integrate them with GBNs to produce a new class of BNs
that have a parametrisation similar to that of CGBNs but make better use of the
information in the data. Under the assumption that each related data set has a
GBN as a generative model, and that all GBNs have the same underlying network
structure but different parameter values, we will show that it is possible to
learn such BNs accurately from data by searching for high-scoring DAGs with
hill-climbing and BIC. Our aim is to complement our previous work
\citep{ijar21-laura} on structure learning from related data sets in discrete
BNs.

This paper is structured as follow. Section~\ref{sec:lmebn} describes linear
mixed-effects models, GBNs and CGBNs; our approach to integrating linear
mixed-effects models with GBNs; and how they can be learned from related data
sets. Section~\ref{sec:simulations} showcases that our approach produces BNs
that outperform GBNs, which disregard the heterogeneous nature of the data, and
CGBNs, which do not perform any pooling, especially at small sample sizes and
when the data are unbalanced. These results are summarised and discussed in
Section~\ref{sec:conclusions}.

\section{Linear Mixed-Effects Models and Bayesian Networks}
\label{sec:lmebn}

Assume that we have an observed variable $F$ denoting which data set each
observation belongs to and that all related datasets comprise the same
continuous variables $\X$. Furthermore, the data must be complete and we assume
that there are no latent variables.

On the one hand, we could model such data with a GBN defined over $\X$,
disregarding their heterogeneous nature, and the local distributions of the
nodes would take the form shown in \mref{eq:gnode}. In this setting, we estimate
the parameters of the GBNs for the related data sets with identical estimates
with a \emph{complete pooling} of their information. This is undesirable because
$F$ will act as a latent variable, possibly introducing spurious arcs in the BN
and biasing the estimates of the $\bXi$. Moreover, the $\sXi$ will be
artificially inflated because the corresponding $\eXi$ are likely to be
heteroscedastic, possibly decreasing the statistical significance of the $\bXi$
and introducing false negative arcs in structure learning.

On the other hand, we could model such data with a CGBN defined over $\X \cup
F$, in which $F$ is the only discrete node and a parent of all the other nodes.
The local distributions of the nodes would take the form shown in
\mref{eq:cgnode} and would be correctly specified. However, each regression in
the mixture would be estimated using only the data of the corresponding data set
and there would be \emph{no pooling} of information across the related data set.
This may make both structure and parameter learning unreliable for sparse data
and for related data sets with unbalanced sample sizes.

\emph{Partial pooling} is a compromise between these two extremes that can
outperform both. We implement it by defining the BN over $\X \cup F$, making $F$
a parent of all the $X_i$ and modelling local distributions by means of
\emph{linear mixed-effects} \citep[LME;][]{pinheiro} models. LME models are
hierarchical models that extend the classic linear regression model by adding a
second set of coefficients, called ``random effects'' and usually denoted
$\bbXi$, which are jointly distributed as a multivariate normal. The other
coefficients are called ``fixed effects''.

Extending the notation of \mref{eq:gnode}, we can write the local distribution
of $X_i$ as
\begin{align}
  &X_i = \mXi + \bPXi\bXi + \Z \bbXi + \eXi,&
    &\bbXi \sim N(\mathbf{0}, \Sigmai), \  \eXi \sim N(\mathbf{0}, \sXi \Im_n),
\label{eq:lmenode}
\end{align}
where $\bPXi$ is the design matrix associated to the nodes in $\X$ that are
parents of $X_i$, $\bXi$ is the vector of fixed effects, $\Z$ is the design
matrix of the random effects and $\bbXi$ is the vector of random effects. For
simplicity, we assume in the following that the sets of random effects
associated with the $|F|$ related data sets are independent of each other and
that we have a random effect for each $\PXi$ and for the intercept. As a result,
$\Z$ is a block matrix of size size $n \times (|\PXi| + 1) |F|$, whose blocks
are associated to different data sets (see \citet{lme4} for more details), and
$\bbXi$ is the $(|\PXi| + 1) |F| \times 1$ vector composed by random effects for
different data sets.

The local distribution in \mref{eq:lmenode} can also be written in the same form
as the local distribution of continuous variable in a CGBN \mref{eq:cgnode}:
\begin{align}
  &\Xij = (\mXd + b_{ij0}) + \bPXi (\bXi + \bbij) + \eXij,
  & \left(
  \begin{array}{c}
    b_{ij0}\\
    \bbij
  \end{array}
  \right) \sim N(\mathbf{0}, \tildeSigmai), \
  \eXij \sim N(\mathbf{0}, \sXi \Im_{n_j}),
\label{eq:lmenode-split}
\end{align}
where $j = 1, \ldots, |F|$ are the related data sets as encoded in the node $F$
which is the only discrete parent $\DXi$; $\bPXi$ from \mref{eq:lmenode}
coincides with $\bGXi$ from \mref{eq:cgnode}; $b_{ij0}$ is the intercept term in
the random effects and the $\bbij$ are the coefficients of the random effects
for the $j$th related data set; $\tildeSigmai$ is the $n_j \times n_j$ block of
$\Sigmai$ associated to the $j$th related data set. The residual variances are
here assumed to be homoscedastic and to be independent of the data set. This
assumption can be easily relaxed to the heteroscedastic case where
$\eXij \sim N(\mathbf{0}, \sXd \Im_{n_j})$.

From \mref{eq:lmenode-split}, we can see that $\bPXi \bbij$ encodes the
deviations of the regression coefficients for the related data sets from the
shared $\bPXi\bXi$ values. Indeed, the coefficients $\bbXi$ associated with the
random effects have mean zero, and  they naturally represent the deviations of
the effects of the parents in individual data sets from their average effects
across data sets, which is represented by the fixed effects $\bXi$. The common
distribution of the random effects $\bbXi$ in \mref{eq:lmenode} produces both a
pooling effect and a shrinkage of the parameters associated to individual data
sets towards their common average. The magnitude of this effect is implicitly
determined by the sample size of each related data set and by the effect size of
each parent, in contrast with the distributional assumptions in
\citet{ijar21-laura} which contain an explicit shrinkage parameter. Moreover,
the inclusion of random effects in \mref{eq:lmenode-split} induces a
decomposition of the variance of $\Xij$ in two independent terms associated with
$\bPXi \bbij$ and $\eXij$, allowing for different variances in different data
sets even though the residuals $\eXij$ are homoscedastic.

We will perform parameter learning using standard statistical results for LMEs,
as described in \citet{demidenko}, while for structure learning we will use BIC
to assess the goodness of fit and implement score-based learning. We will not
consider constraint-based learning, which can be implemented using statistical
tests for nested models to simultaneously test the significance of the fixed and
random effects associated with a parent, following \citet{pinheiro} and
\citet{demidenko}. The expression for the BIC score arising from
\mref{eq:lmenode} has the same general form as \mref{eq:bic}, but the
characterisation of number of parameters $|\T|$, where $\T = \{ \bXi, \bbXi,
\Sigmai, \sXi\}$, is more complex. In a GBN, each local distribution has $|\PXi|
+ 2$ parameters; in a CGBN, $(|\PXi| + 2) |\DXi|$. In the proposed model we
cannot simply add the number of random effects to the count: we must also
estimate their covariance structure $\Sigmai$.
Overall, the number of free parameters is
\begin{equation}
  |\T| = \underbrace{
           \vphantom{{|\PXi| + 1 \choose 2}}
           |\PXi| + 1
         }_{\text{fixed effects}} +
         \underbrace{
           |\PXi| + 1 + {|\PXi| + 1 \choose 2}
         }_{\text{random effects}} +
         \underbrace{
           \vphantom{{|\PXi| + 1 \choose 2}}
           1
         }_{\text{residuals}}
       = \frac{|\PXi|^2 + 5 |\PXi| + 6}{2}\, ,
\label{eq:nparams}
\end{equation}
since $\Sigmai$ contains $|\PXi| + 1$ variances and ${|\PXi| + 1 \choose 2}$
covariances, if there are $|\PXi| + 1$ random effects.

Although it is beyond the scope of this paper, note that we can leverage the
similarities between \mref{eq:lmenode} and \mref{eq:cgnode} to improve the
learning of general CGBNs by using generalised LMEs \citep{demidenko} and thus
inducing information pooling and shrinking in their local distributions.

\section{Simulation Study}
\label{sec:simulations}

We showcase the properties of this combination of LMEs and BNs with a simulation
study. The experimental design is as follows:
\begin{enumerate}
  \item For each of $N = 10, 20, 50$, for each of $\AP = 1, 2, 4$, and for each
    $|F| = 2, 5, 10, 20, 50$ we generate DAGs ensuring that all nodes are
    connected, and that there is an arc pointing from $F$ to each node. Here
    $\AP$ denotes the average number of parents of each node $X_i$ and
    represents the density of the DAG. In each DAG, arcs are included
    independently of each other with probability $p = \AP \cdot 2 / N$ to obtain
    $\AP$ arcs on average, and DAGs that are not connected are discarded.
  \item For each DAG, we assign $F$ a uniform
    distribution to simulate a balanced sample. For each of the continuous
    variables $X_i$ and each related data set $j = 1, \ldots, |F|$, we sample
    the regression coefficients of \mref{eq:cgnode} as
    \begin{align*}
      &\bXd \sim N(\bXi + \bbij, \sigma^2_{\bXd} \Im_{|\PXi|+1}),&
      &\bXi = \mathbf{2}, \,
       \bbij \sim N(\mathbf{0}, \Im_{|\PXi|+1}), \sigma^2_{\bXd} \sim \chi^2_1&
    \end{align*}
    and we set the standard error of the residuals $\sXd$ so that the $\PXi$
    explain 85\% of the variance of $X_i$ for each of the $|F|$ related data
    sets, thus ensuring local distributions are not singular.
\end{enumerate}
We generate 5 BNs (denoted $\BT$) for each configuration of $N$, $\AP$, $|F|$
and for each BN we generate different balanced and unbalanced data sets with
different number of observations.

\begin{figure}
  \includegraphics[width=\linewidth]{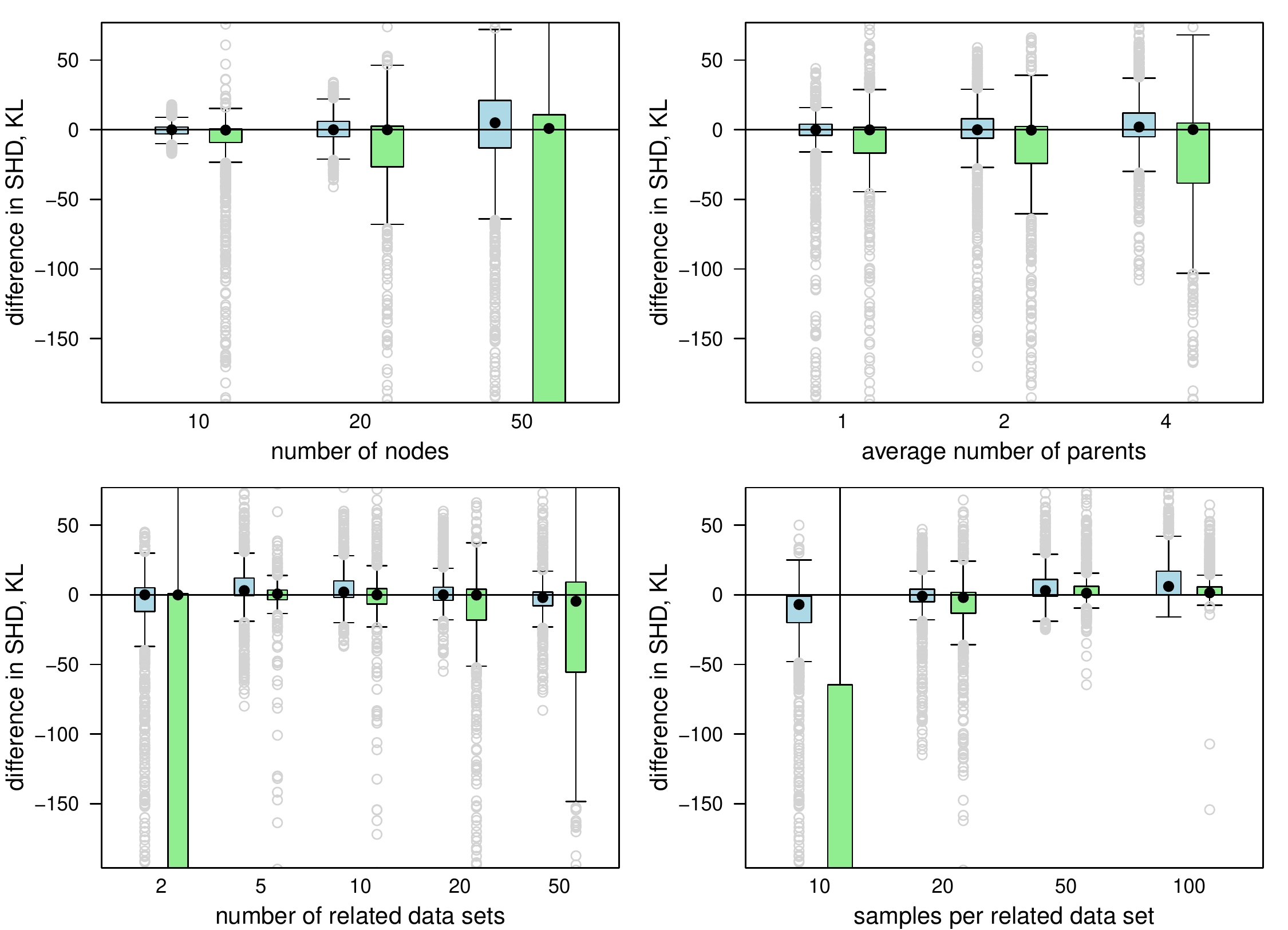}
  \caption{$\SHD(\BLME) - \SHD(\BCG)$ (blue) and
    $\KL(\BT, \BLME) - \KL(\BT, \BCG)$ (green) for all values of $N$ (top left),
    $\AP$ (top right), $|F|$ (bottom left) and $n_j$ (bottom right). Negative
    values favour $\BLME$. The range of the KL differences is such that it is
    impossible to show the whole extent of some boxplots while keeping the
    others visible.}
\label{fig:balanced}
  \includegraphics[width=\linewidth]{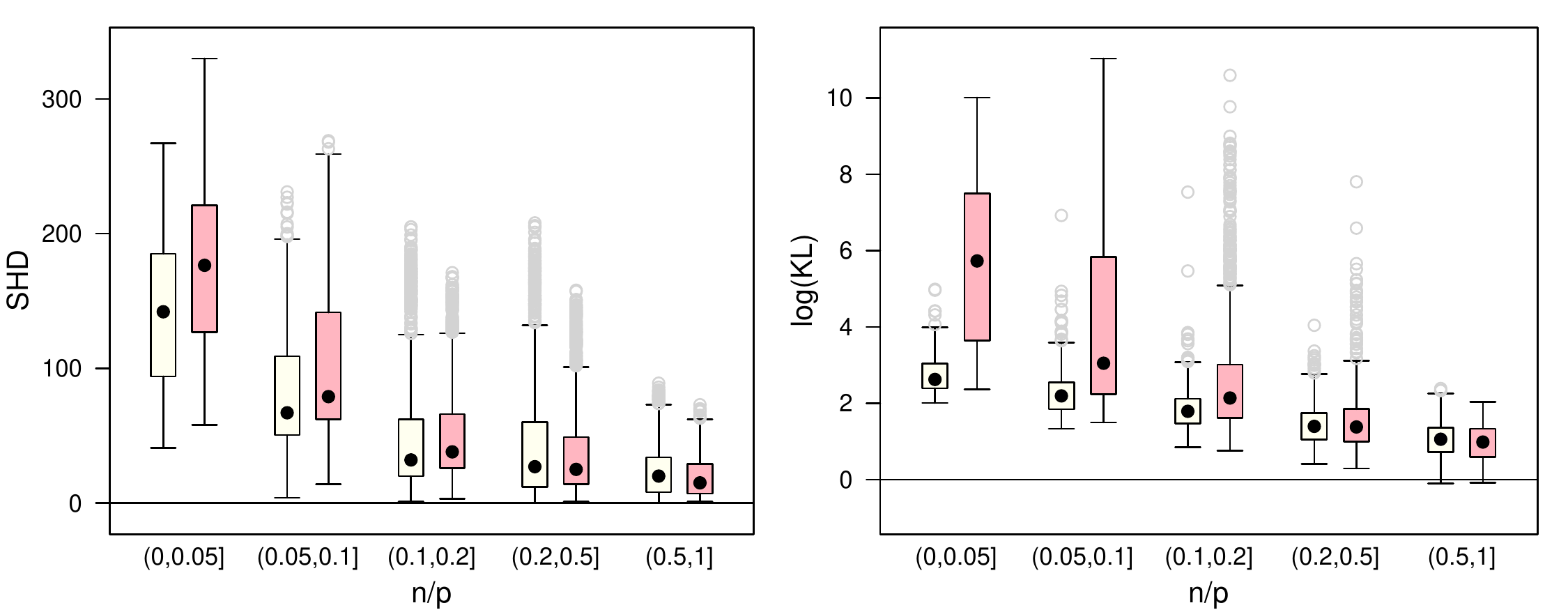}
  \caption{SHD (left) and KL (right) for the $\BLME$ (ivory) and the $\BCG$
    (pink).}
\label{fig:efficiency}
\end{figure}

From each data set, we learn three different types of BNs using the
implementation of hill-climbing and BIC in the \textbf{bnlearn} R package
\citep{jss09}:
\begin{enumerate}
  \item \emph{Complete pooling} ($\BG$): we learn a GBN from $\X$, completely
    disregarding $F$.
  \item \emph{No pooling} ($\BCG$): we learn a CGBN from $\{\X, F\}$ following
    \mref{eq:cgnode}.
  \item \emph{Partial pooling} ($\BLME$): we learn a CGBN from $\{\X, F\}$
    following \mref{eq:lmenode}.
\end{enumerate}
We assess the structural accuracy of the learned DAGs with the Structural
Hamming distance \cite[SHD;][]{mmhc} between the network structure of $\BT$ and
each of the structures of the $\BG$, $\BCG$, $\BLME$ learned from the data
generated from $\BT$. Similarly, we asses the parametric accuracy of the BNs by
computing the Kullback-Leibler distance \cite[KL;][]{kullback} between $\BT$
(with the true parameter values) and $\BG$, $\BCG$, $\BLME$ (where parameters
are estimated from data). For brevity, we will only discuss how these metrics
compare for $\BCG$ and $\BLME$: the $\BLME$ have lower SHD values than the $\BG$
for 94\% of the data sets in our simulation and lower KL for 96\%. Even in the
most adverse combination of experimental settings ($N = 10$, $\AP = 4$), the
$\BLME$ have lower SHD for 65\% of the data sets and lower KL for 93\%. With
such a stark difference in performance, it is clear that $\BLME$ should be
always be preferred to $\BG$.

\subsection{Balanced Data Sets}

Firstly, we generate 5 data sets with $n_j = 10, 20, 50, 100$ observations in
each related data set from each generated BN. Figure~\ref{fig:balanced} shows
the difference between the $\BLME$ and the $\BCG$ in terms of SHD and KL. The
$\BLME$ increasingly outperform the $\BCG$ in terms of KL as the number of nodes
increases (top left), as the density of the networks increases (top right) and
as the number of related data sets grows from 5 to 50 (bottom left).
Unexpectedly, the $\BLME$ markedly outperform the $\BCG$ when $|F| = 2$ even
though mixed-effects models are known to provide better performance for large
$|F|$ \citep{pinheiro}. The two approaches appear to be tied in terms of SHD,
and to have similar rates of false negative arcs (that are in $\BT$ but not in
$\BCG$ or $\BLME$) and of false positive arcs (that are in $\BCG$ or $\BLME$ but
not in $\BT$). However, we can see in the bottom-right panel that for $n_j = 10$
the $\BLME$ outperform the $\BCG$ also in terms of SHD. As $n_j$ increases, both
the $\BLME$ and the $\BCG$ gradually converge to their large sample behaviour
and learn the generating BN well, with only minor differences in both SHD and
KL.

This leads us to investigate the relative sample efficiency of the $\BLME$ and
the $\BCG$ in Figure~\ref{fig:efficiency}. Plotting both SHD (left panel) and KL
(right panel) against the number of samples per parameter in $\BT$ (denoted $n/p$
with $n = |F|n_j$ and $p = |\Theta_{\BT}|$), we can see  that the $\BLME$ have
lower SHD and lower KL than the $\BCG$ for ratios up to 0.2. In particular, the
difference in KL can easily be an order of magnitude in that range. We attribute
this dominance to the pooling effect, which can result in better network scoring
and better parameter estimation at lower sample sizes when sharing information
has the most impact. When the the number of samples per parameter increases
above 0.5 the two approaches are roughly equivalent, because each related data
set contains enough information to score networks and estimate parameters
appropriately and pooling has therefore a smaller effect. Notably the $\BCG$ do
not outperform the $\BLME$ even in this case.

\subsection{Predictive and Classification Accuracy}

\begin{figure}[t]
  \includegraphics[width=\linewidth]{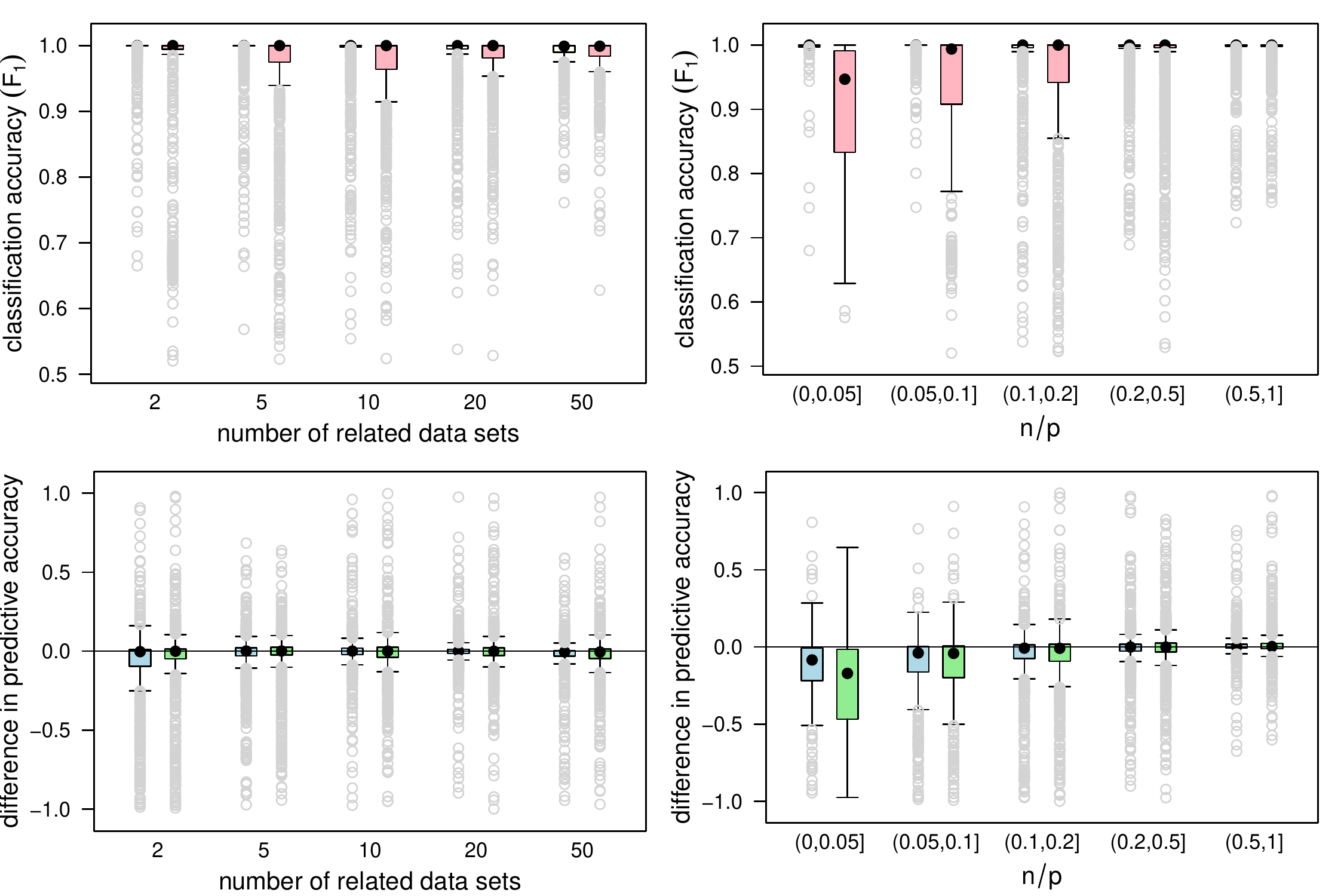}
  \caption{Classification accuracy against $|F|$ (top left) and $n/p$ (top
    right) for the $\BLME$ (ivory) and the $\BCG$ (pink) for balanced data sets.
    Difference in predictive accuracy between the $\BLME$ and the $\BCG$ against
    $|F|$ (bottom left) and $n/p$ (bottom right) knowing $F$ (blue) and not
    knowing $F$ (green); negative values favour $\BLME$.}
\label{fig:predictive-balanced}
\end{figure}

We also evaluate the $\BLME$ and the $\BCG$ on their:
\begin{itemize}
  \item \emph{Predictive accuracy:} predicting each $X_i$ from other variables
    in $\X$, with and without knowledge of $F$. Measured with the average over
    the $X_i$ of the relative errors between observed and predicted values (here
    denoted $x_{ik}$ and $\widehat{x}_{ik}$ for node $X_i$ and observation $k$)
    in absolute value:
    \begin{equation*}
      \RMAD = \frac{1}{N} \sum_{i = 1}^N \left[\,
        \frac{1}{n} \sum_{k = 1}^n
          \left|\frac{x_{ik} - \widehat{x}_{ik}}{x_{ik}}\right|
      \,\right]
    \end{equation*}
  \item \emph{Classification accuracy:} predicting $F$ from $\X$. Measured with
    the \Fone{} score (the harmonic average between precision and recall).
    Following \citet{kuhn}, when $|F|$ is greater than 2 we compute the \Fone{}
    score as the average of the one-vs-rest \Fone{} scores for each related data
    set.
\end{itemize}
To evaluate both measures of accuracy, we generate an additional sample of 1000
observations from $\BT$ for each BN in our experimental design. Predicted values
are computed as the posterior expectations for the relevant variables using
likelihood-weighting approximate inference \citep{koller}. We show both
performance measures in Figure~\ref{fig:predictive-balanced}. From the top-left
plot, we can see that the $\BLME$ always have larger \Fone{} values than the
$\BCG$, and that they have only negligible departures from $\text{\Fone} = 1$ up
to $|F| = 20$. We can also see that the $\BLME$ dominate the $\BCG$ for all
$n/p$, and that the $\BCG$ only approach the $\BLME$ in performance as $n$
approaches $p$ top right panel). Similarly, the difference in predictive
accuracy between the $\BLME$ and the $\BCG$ is always in favour of the former
until $n/p < 0.2$ for both known and unknown $F$ (bottom right panel), and it is
not markedly different from zero for all the considered $|F|$ (bottom left
panel).

\begin{figure}[t]
 \includegraphics[width=\linewidth]{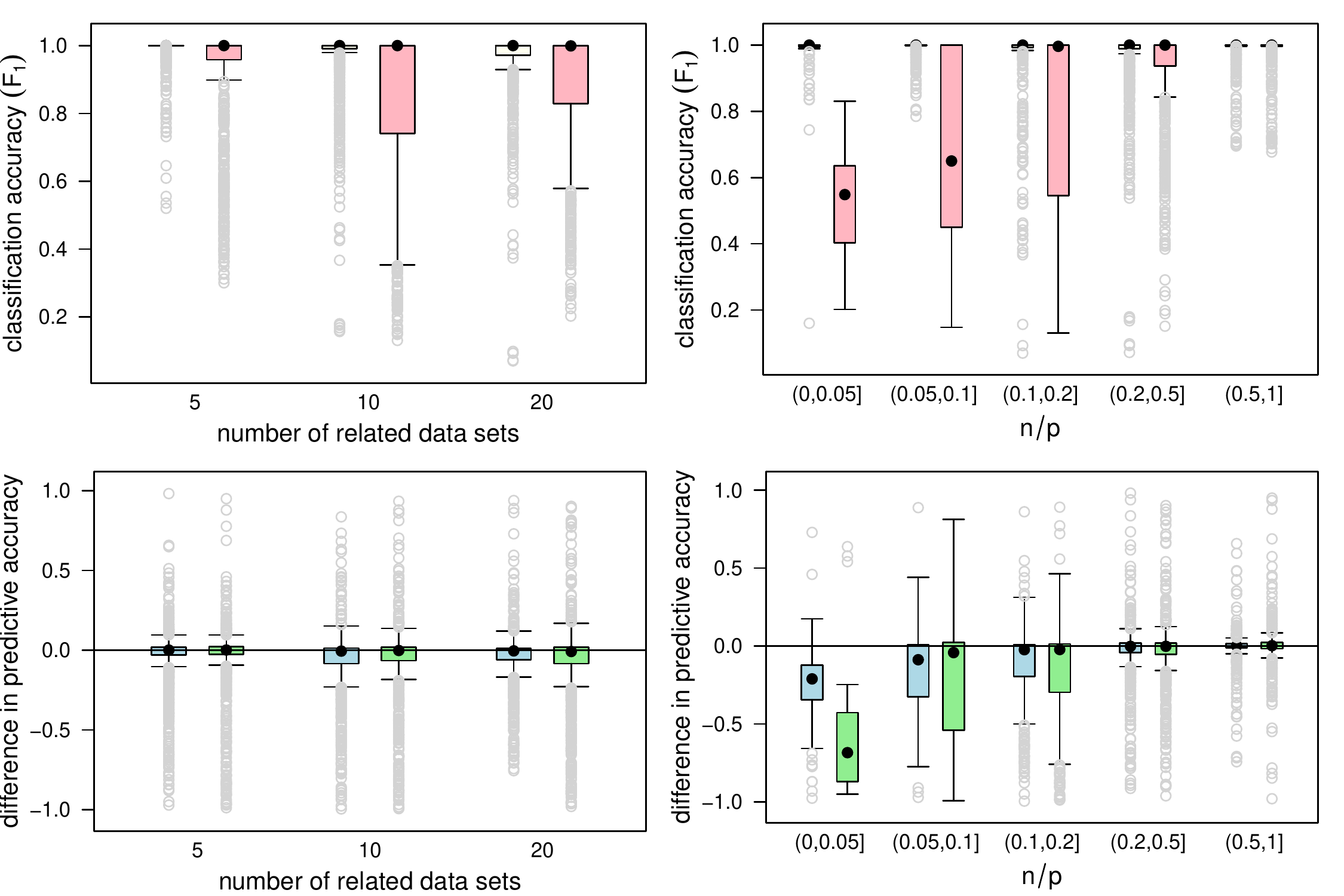}
  \caption{Classification accuracy against $|F|$ (top left) and $n/p$ (top
    right) for the $\BLME$ (ivory) and the $\BCG$ (pink) for unbalanced data
    sets. Difference in predictive accuracy between the $\BLME$ and the $\BCG$
    against $|F|$ (bottom left) and $n/p$ (bottom right) knowing $F$ (blue) and
    not knowing $F$ (green); negative values favour $\BLME$.}
\label{fig:predictive-unbalanced}
\end{figure}

\subsection{Unbalanced Data Sets}

Another scenario in which pooling can make a difference is when the sample is
unbalanced, that is, when some of the related data sets contain markedly fewer
observations that others. To evaluate the effect of the lack of balance, we
modify our simulation experiments as follows:
\begin{enumerate}
  \item we consider only those $\BT$ with $|F| = 5, 10, 20$;
  \item for each of them, we set the overall sample size to $n = |F|n_j$,
    with $n_j = 10, 20, 50, 100$;
  \item and we assign $0.3n$ observations to each of two related data sets while
    distributing the remaining $0.4n$ evenly among the others.
\end{enumerate}

The resulting $\BLME$ and $\BCG$ behave largely as with balanced data sets, but
with $\BLME$ increasingly outperforming $\BCG$ as $|F|$ increases (and, as a
result, the data sets become more and more unbalanced). For instance, the
proportion of the simulations in which $\BLME$ outperforms $\BCG$ in terms of KL
changes from 52\% (balanced data sets) to 60\% (unbalanced data sets) when $|F|
= 20$. Similarly, $\BLME$ increasingly outperforms $\BCG$ as $n/p$ decreases,
with $\BLME$ outperforming $\BCG$ in 78\% of the simulations (unbalanced data
set) instead of 75\% (balanced data sets) in which $n/p < 0.2$. Plotting the KL
and SHD of $\BLME$ and $\BCG$ side by side produces boxplots that are very
similar to those in Figure~\ref{fig:efficiency} both marginally and
conditionally on the individual values of $|F|$ (figures not shown for brevity).
The same is true for predictive and classification accuracies, which we show in
Figure~\ref{fig:predictive-unbalanced}. The difference in classification
accuracy is increasingly in favour of $\BLME$ as $|F|$ increases (top left) and
as $n/p$ decreases (top right). On the other hand, predictive accuracy is
similar for all $|F|$ (bottom left), but is increasingly in favour of $\BLME$ as
$n/p$ decreases for both known and unknown $F$ (bottom right). The improvement
in both the predictive and classification accuracy of $\BLME$ is more marked in
the unbalanced case compared to the balanced one.

\subsection{Homogeneous Data Sets}

\begin{figure}[t]
 \includegraphics[width=\linewidth]{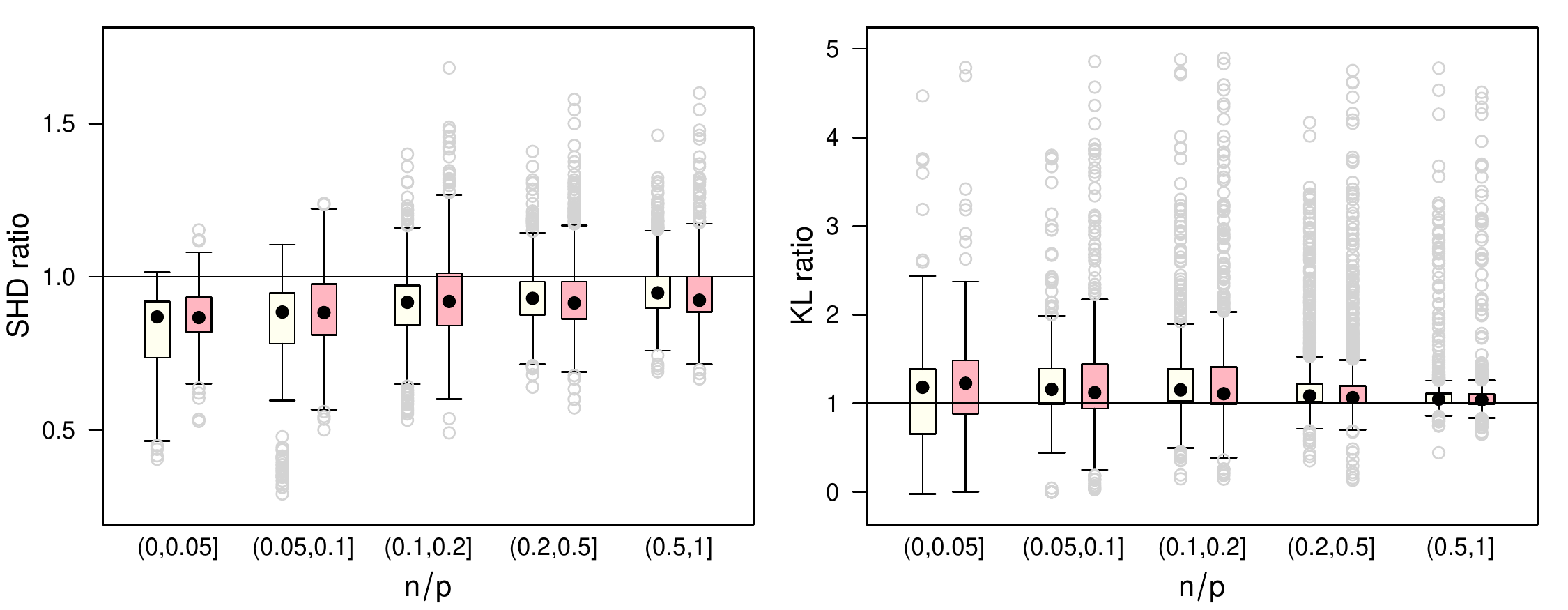}
  \caption{Left panel: $\SHD(\BLME) / \SHD(\BG)$ (ivory) and
    $\SHD(\BCG) / \SHD(\BG)$ (pink). Right panel:
    $\KL(\BT,\BLME) / \KL(\BT,\BG)$ (ivory) and $\KL(\BT,\BCG) / \KL(\BT,\BG)$
    (pink). Both are computed from homogeneous data and shown as a function of
    $n/p$. Values larger than 1 favour $\BG$.}
\label{fig:homogeneous}
\end{figure}

Finally, we compare $\BLME$, $\BG$ and $\BCG$ when the data are homogeneous,
that is, when the data do not comprise multiple related data sets with different
distributions. For this purpose, we take the $\BT$ we generated earlier and we
modify them so that
\begin{align*}
  &\bXd = \boldsymbol{\beta}_{i1}& &\text{and}& &\sXd = \sigma^2_{i1}&
  &\text{for all $i = 1, \ldots, N$ and $j = 1, \ldots, |F|$.}
\end{align*}
We then proceed to resample 5 balanced data sets with $n_j = 10, 20, 50, 100$ as
in the first simulation setup (balanced data sets). Note that $\BT$ still
contains a node $F$ identifying  $|F| = 2, 5, 10, 20, 50$ related data
sets, so the generated data will contain a set of labels that are effectively
non-informative of the distribution of $\X$. In this scenario, $\BG$ is the
correctly specified model while both $\BLME$ and $\BCG$ are over-parametrised.
For this reason we will compare $\BLME$ and $\BCG$ relative to $\BG$. Relevant
results from this modified simulation are shown in Figure~\ref{fig:homogeneous}.

In terms of SHD, $\BLME$ is roughly equivalent to $\BCG$ (better 35\% of the
time, equal 16\%, worse 47\%): computing $\SHD(\BLME) / \SHD(\BCG)$ we find that
the median ratio is equal to $1$ and the interquartile range is $[0.95, 1.05]$,
that is, 50\% of the simulations fall in this interval (figures are not shown
for brevity). Furthermore, $\BLME$ is better than $\BG$ 75\% of the time but the
improvement is modest: the ratio $\SHD(\BLME) / \SHD(\BG)$ has median $0.93$ and
interquartile range $[0.87, 1]$. We attribute this effect to the implicit
regularisation provided by LMEs over classical regressions model as a result of
shrinking the coefficients $\bbij$.

In terms of KL, we obtain as expected that $\KL(\BT, \BG)$ is smaller than
$\KL(\BT, \BLME)$ for 78\% of the simulations and smaller than $\KL(\BT, \BCG)$
for 75\% of the simulations. The difference in performance, however, is small:
$\KL(\BT, \BLME) / \KL(\BT, \BG)$ has median $1.07$ and interquartile range $[1,
1.23]$, and $\KL(\BT, \BCG) / \KL(\BT, \BG)$ has median $1.07$ and interquartile
range $[1, 1.35]$. This suggests that even though  $\BCG$ is more
over-parametrised than $\BLME$, the impact of over-parametrisation is modest.
And in fact, we find that $\BLME$ and $\BCG$ are very close in performance: the
ratio $\KL(\BT, \BLME) / \KL(\BT, \BCG)$ has median $1.001$ and interquartile
range $[0.923, 1.027]$ (figures not shown for brevity).

If we restrict ourselves to simulations for which $n / p < 0.1$, the difference
between $\BLME$ and $\BCG$ becomes more marked in terms of KL
($\KL(\BT, \BLME) / \KL(\BT, \BCG)$ decreases in median from $1.001$ to $0.731$
and the interquartile range shifts to $[0.003, 1.047]$), while it remains
comparable over the complete set of simulations in terms of SHD \linebreak
($\SHD(\BLME) / \SHD(\BCG)$ has median $1.01$ and interquartile range  $[0.88,
1.07]$). Hence we conclude that the $\BLME$ fit homogeneous data better than the
$\BCG$ for small sample sizes. The difference between the two becomes smaller as
the sample size grows and the pooling effect is reduced, as expected from the
theoretical foundations of LMEs in Section \ref{sec:lmebn}.

As for predictive accuracy, $\BLME$ has lower error rates than $\BCG$ for 60\%
of the simulations but the difference between the two is small both for the
whole set of simulations and when we restrict ourselves to simulation for which
$n / p < 0.1$ (figures not shown for brevity). We attribute this effect to the
implicit regularisation introduced by LMEs. The ratio $\RMAD(\BLME) /
\RMAD(\BCG)$ has median $1.03$ in the former case and $1.1$ in the latter,
although the interquartile range increases in size from $[0.942, 1.306]$ to
$[0.818, 2.190]$ suggesting more variability for small sample sizes. The
comparison of $\BG$ with respect to $\BLME$ and $\BCG$ shows that $\BG$ performs
at least as well as $\BLME$ and $\BCG$ in almost all simulations (figures not
shown for brevity). Moreover, all the results in terms of predictive accuracy
are equivalent when $F$ is known and unknown, since $\X$ is now independent from
$F$ by construction. For the same reason, classification accuracy is not
relevant in this simulation scenario.

\section{Conclusions}
\label{sec:conclusions}

In this paper we revisited the problem of learning the structure of a BN from
related data sets. In our previous work \citep{ijar21-laura} we focused on
discrete BNs. Here we consider GBNs instead: we propose to use LMEs to produce a
CGBN in which local distributions are replaced with LMEs to separate the effects
that are specific to individual data sets (the random effects) from those that
are common to all of them (the fixed effects).

LMEs are well-established models in the literature, and their favourable
properties are advantageous in BN learning. In particular, the automatic pooling
of information between the related data sets results in BNs with better
structural and parametric accuracy for small sample sizes and unbalanced data
sets (that is, when some of the related data sets are markedly smaller than
others). Our experimental evaluation suggests that our approach always
outperforms GBNs because they disregard the heterogeneity in the data (complete
pooling). Our approach is also at least as good as using a standard CGBN (no
pooling) in terms of structural accuracy, and outperforms it in terms of
parametric accuracy in most simulations. Even when the data are not
heterogeneous, our approach has comparable performance despite being
over-parametrised compared to a correctly specified (GBN) model.

Furthermore, our approach can be extended to model related data sets containing
both discrete and continuous variables by using generalised LMEs. The
performance observed in this paper is encouraging in this respect.


\end{document}